\documentclass[10pt,twocolumn,letterpaper]{article}

\usepackage[accsupp]{axessibility} % Improves PDF readability for those with disabilities.
\usepackage{iccv}
\usepackage{times}
\usepackage{epsfig}
\usepackage{graphicx}
\usepackage{amsmath}
\usepackage{amssymb}
\usepackage{multirow}
\usepackage{comment}

\usepackage[skip=2pt]{caption} % example skip set to 2pt
\usepackage{subcaption}
\usepackage{hhline}

\usepackage[pagebackref=true,breaklinks=true,letterpaper=true,colorlinks,bookmarks=false]{hyperref}

\iccvfinalcopy % *** Uncomment this line for the final submission

\def\iccvPaperID{****} % *** Enter the ICCV Paper ID here

\ificcvfinal\fi

\begin{document}

%%%%%%%%% TITLE
\title{Sparse Linear Concept Discovery Models}

\begin{comment}
\author{Konstantinos P. Panousis\\
Inria\\
{\tt\small konstantinos.panousis@inria.fr}
\and
Diego Marcos\\
Inria\\
{\tt\small diego.marcos@inria.fr}
\and
Dino Ienco\\
INRAE/UMR TETIS/Univ. of Montpellier\\
{\tt\small dino.ienco@inrae.fr}
}
\end{comment}
\author{Konstantinos P. Panousis$^{1,3,4,5}$  \quad Dino Ienco$^{2,3,5}$ \quad Diego Marcos$^{1,3,4,5}$
\\ \\
 $^{1}$Inria \quad $^{2}$Inrae \quad $^{3}$University of Montpellier \quad $^{4}$LIRMM \quad $^{5}$UMR-Tetis  \\
{\tt\small \{konstantinos.panousis@inria.fr, diego.marcos\}@inria.fr}
\quad
{\tt\small dino.ienco@inrae.fr}
}

\maketitle
% Remove page # from the first page of camera-ready.
\ificcvfinal\thispagestyle{empty}\fi

\begin{abstract}
    The recent mass adoption of DNNs, even in safety-critical scenarios, has shifted the focus of the research community towards the creation of inherently intrepretable models. Concept Bottleneck Models (CBMs) constitute a popular approach where hidden layers are tied to human understandable concepts allowing for investigation and correction of the network's decisions. However, CBMs usually suffer from: (i) performance degradation and (ii) lower interpretability than intended due to the sheer amount of concepts contributing to each decision. In this work, we propose a simple yet highly intuitive interpretable framework based on Contrastive Language Image models and a single sparse linear layer. In stark contrast to related approaches, the sparsity in our framework is achieved via principled Bayesian arguments by inferring concept presence via a data-driven Bernoulli distribution. As we experimentally show, our framework not only outperforms recent CBM approaches accuracy-wise, but it also yields high per example concept sparsity, facilitating the individual investigation of the emerging concepts. Our code and models are available at: \url{https://github.com/konpanousis/ConceptDiscoveryModels}.
\end{abstract}

\section{Introduction}
Deep Neural Networks (DNNs) have been established as the de-facto state-of-the-art approach for a variety of domains and applications. Their performance has facilitated their widespread adoption, especially in CV and NLP, including safety-critical tasks such as autonomous driving and healthcare. However, due to their highly complex structure, DNNs are considered \textit{black-box} models: they map an input to an output via an \textit{un-interpretable} computation process. This constitutes a highly undesirable property, especially in safety- or bias-aware domains, where trustworthiness via the interpretation of the decision making process is key. Thus, conceiving \textit{inherently} interpretable networks constitutes a crucial research and societal challenge. 

One of the best known approaches in this context, is Concept Bottleneck Models (CBMs) \cite{koh20a}. CBMs commonly comprise two basic structures: (i) a Concept Bottleneck Layer (CBL) trained to tie its neurons to human interpretable \textit{concepts}, e.g., textual descriptions, followed by (ii) a \textit{linear} decision layer that facilitates the interpretability of the decision process since it is now based on an affine combination of the learned concepts. Despite this more interpretable mode of operation, CBMs suffer from three significant drawbacks: (i) need for labeled data for the predefined concepts, (ii) performance degradation compared to a standard neural backbone, and (iii) rely on \textit{implicit} interpretation of the contribution of each concept, to the final decision, through the analysis of the last linear layer weights. 

In this work, we aim to address the limitations of current CBMs by introducing a novel framework for interpretable neural networks based on: (i) recent advances in CLIP-based models, (ii) a single linear decision layer, and (iii) a novel per example \textit{explicit} concept discovery and sparsification mechanism that builds upon solid Variational Bayesian arguments. We dub our approach Concept Discovery Models (CDMs). As we experimentally show, our framework significantly outperforms recent CBM-based SOTA alternatives accuracy-wise, while giving rise to a principled data-driven mechanism for discovering a \textit{highly flexible} and \textit{highly sparse} set of concepts for each example.

\section{Related Work}
% \paragraph{Contrastive Language Image Pretraining.}
% CLIP \cite{radford21a} models leverage natural language supervision for image representation learning, relying on (image, text) pairs and jointly training an image and a text encoder. These encoders are trained to maximize the cosine similarity between the real pairs while minimizing the similarity for all other possible pairings. During inference, we can leverage the learned encoders in order to predict how similar (in the cosine similarity sense) a specific test image and text snippet are. 

\paragraph{Concept Bottleneck Models \& Sparsity.}
The most similar approach to ours is Concept Bottleneck Models (CBMs)~\cite{koh20a}. CBMs have facilitated recent developments towards interpretable architectures, with many methods aiming to alleviate their drawbacks, e.g., performance degradation. Post-hoc CBMs \cite{yuksekgonul2022posthoc} constitute such an extension: any backbone is made interpretable through training a single FC layer, while optionally performing residual fitting to restore any performance loss. More recently, Label-Free (LF) CBM~\cite{labelfree} was introduced, an ``automated'' CBM with a sparse linear prediction layer, in which four steps are considered: (i) automatic concept creation using GPT and filtering, (ii, iii) computation of the CBL projection weights through the CLIP-based concept similarity matrix~\cite{radford21a}, and (iv) training the sparse final layer. Despite these advances, most works~\cite{wong2021,labelfree,marcos2020contextual} resort to complicated schemes for separately training the CBL and the linear layer, relying on impromptu constraints that may harm interpretability. Specifically, the sparse layer is trained post-hoc, using custom solvers that require ad-hoc application-specific sparsity or accuracy thresholds, despite the existence of more data-driven approaches in the literature. 

In this context, recent advances in Variational Bayesian methods towards pruning or component omission \cite{chatzis2018indian, panousis2019nonparametric, panousis2022competing} have paved the way for a principled data-driven and end-to-end trainable sparsity mechanism: auxiliary binary latent variables are introduced to explicitly model the presence or absence of network components in an ``on''-``off'' fashion, without requiring any ad-hoc thresholds. We exploit this rationale and construct a principled framework that explicitly \textit{infers} the per-example concept presences, allowing for varying and unrestricted flexibility.

\section{Proposed Approach}
% In this work, we build upon recent CLIP advances and devise a novel framework towards interpretability. We construct a CBM, by training a single linear layer that ties its neurons to concepts, i.e., textual descriptions, by leveraging the similarities between images and concepts arising from CLIP. At the same time, the considered linear layer is sparsified in a data-driven way by exploiting principled Variational Bayesian arguments as we describe next.

Let us denote by $\boldsymbol X\in\mathbb{R}^{N\times H \times L \times c}$, a dataset comprising $N$ images, each with height $H$, width $L$ and $c$ channels, and by $\boldsymbol A=\{a_1, a_2, \dots, a_{M}\}$ a predefined set of attributes/concepts, where $M = |\boldsymbol A|$ denotes the dimensionality of the set, i.e., the number of concepts. 

Image-Language models, e.g., CLIP~\cite{radford21a}, typically comprise an image encoder $E_I(\cdot)$ and a text encoder $E_T(\cdot)$; these are jointly trained in a contrastive manner to learn a \textit{common embedding space}~\cite{SimCLR,npairs_loss}. During inference, we first project images and text to this common space; therein, we can compute their similarity using these ($\ell_2-$normalized) \textit{embeddings}. The cosine similarity is usually considered, computed via the inner product:
\begin{equation}    
    \label{eqn:cos_sim}
    \resizebox{.9\linewidth}{!}{$%
    \mathrm{Cos \ Similarity}\triangleq \boldsymbol S \propto E_I (\boldsymbol X) E_T(\boldsymbol A)^T \in \mathbb{R}^{N \times M }    
    $}
\end{equation}
 Since $\boldsymbol S$ is computed between all possible images-concepts pairings, it yields a unique representation for each image, encoded via the similarity with each distinct concept; thus, these per-image representations can be naturally employed to support a downstream task.

%it is instictive to employ these per-image representations directly to achieve a downstream task. 

We consider classification using a single linear layer comprising a weight matrix $\boldsymbol W_c \in \mathbb{R}^{C \times M}$, where $C$ is the number of classes. During training and since we use the similarity value as an input, %this matrix 
$\boldsymbol W_c$ will learn to encode how each concept relates to each particular class. The output of the network $\boldsymbol{Y}\in\mathbb{R}^{N \times C}$ yields:
\begin{equation}
    \label{eqn:simplest_form}
    \resizebox{.73\linewidth}{!}{$%
    \boldsymbol Y = \boldsymbol S \boldsymbol W_c^T 
    \propto \left(E_I(\boldsymbol X) E_T(\boldsymbol A)^T \right) \boldsymbol W_c^T
    $}
\end{equation}

\begin{figure*}
    \centering
    \begin{subfigure}[c]{0.4\textwidth}
        \centering
        \includegraphics[scale = 0.35]{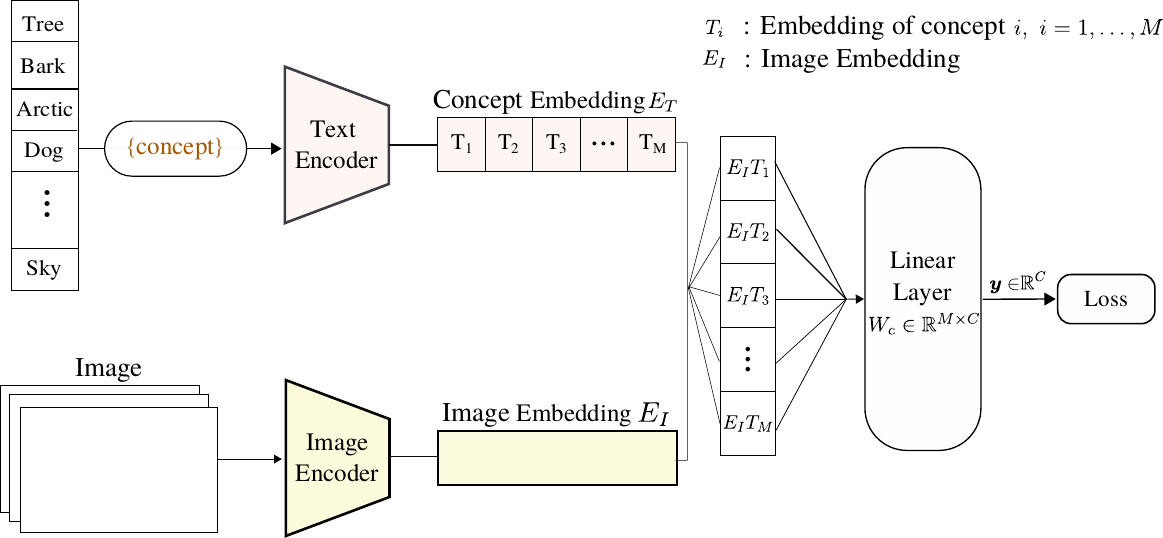}
    \label{fig:concept_simple}
    \end{subfigure}\hfill
    \begin{subfigure}[c]{.6\textwidth}
        \centering
        \includegraphics[scale = 0.4]{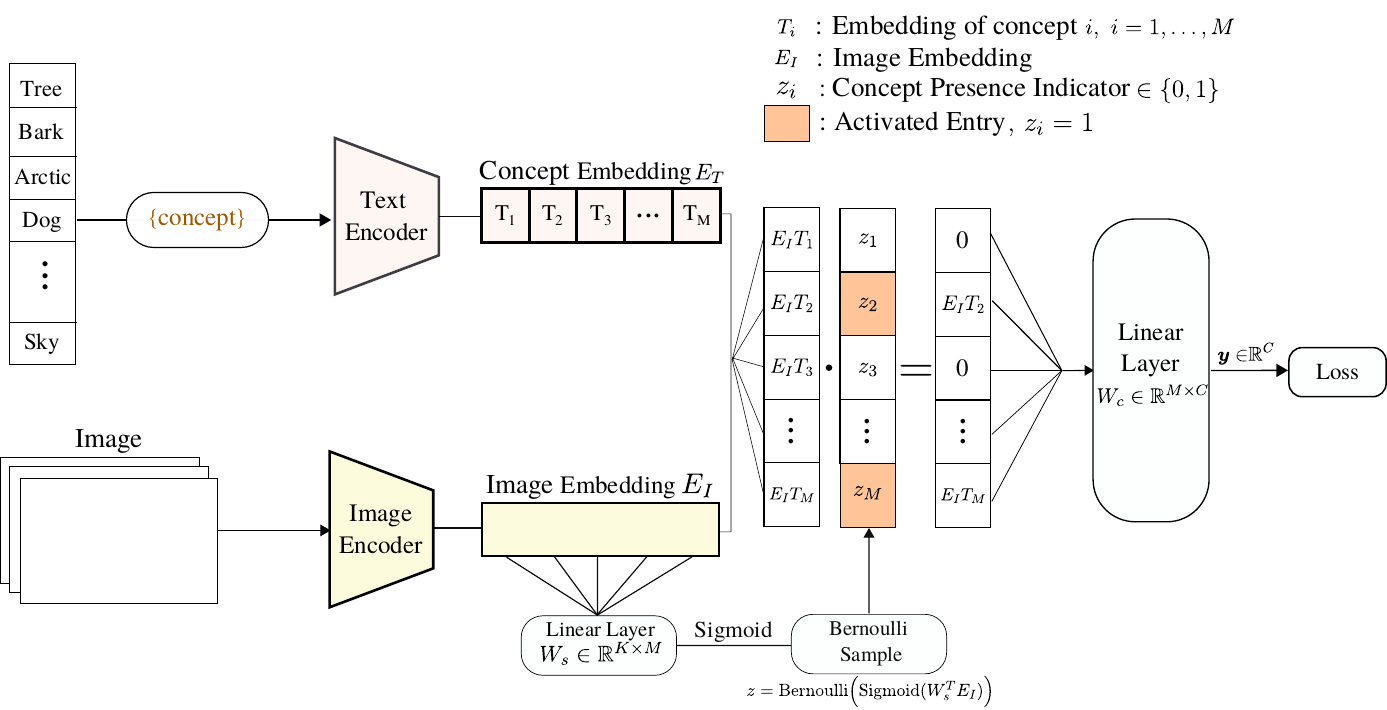}
        \label{fig:concept_bernoulli}
    \end{subfigure}
    \caption{(Left) The base model utilizes the similarities between the images-concepts CLIP embeddings to perform classification with a single linear layer. (Right) The proposed data-driven concept discovery framework. In this case we exploit the information of the image embeddings to devise a mechanism for explicit concept presence indication per example.}
    \label{fig:archs}
    \vspace{-.5em}
\end{figure*}

In  Eq. \eqref{eqn:simplest_form}, we \textit{linearly} combine all concept-related information and compute a class probability \textit{for each image}. %In stark contrast 
Conversely to related CBMs approaches that rely on complicated projections, we posit that the CLIP similarity vector presents a sufficient images-concepts representation without the need for any additional computational overheads.  We then use the standard cross-entropy loss for classification. A graphical illustration is depicted in Fig. \ref{fig:archs} (Left).

However, the commonly used linear decision layer has a significant drawback: the relation between images and concepts is \textit{implicit}. Indeed, most approaches rely on the magnitudes of $\boldsymbol{W}_c$ and the projections to assess the effect of \textit{all} concepts, without considering information redundancy. Recent approaches \cite{wong2021,labelfree}, aim to alleviate this issue by sparsifying the linear layer \textit{for each class}, leveraging however complicated solvers that require tuning of an ad-hoc task specific cut-off thresholds. We posit that restricting the concepts per class to a fixed set, greatly limits the flexibility of appropriate per-example concept representation.

To bypass these limitations, and in stark contrast to other sparsity-inducing approaches, we propose a novel, data-driven formulation for inferring the essential number of concepts present on a per-example basis. To this end, we introduce a set of auxiliary\textit{ binary latent indicators} $\boldsymbol Z \in \{0,1\}^{N \times M}$; these denote whether a particular concept considered for each example in an ``on''-``off'' fashion; that is, $z_{n,m}=1$ if concept $m$ is active for example $n$, $z_{n,m}=0$ otherwise. Thus, instead of only relying on \textit{implicit} measures, we have now defined an \textit{explicit} mechanism of concept presence. The output  (Eq.\eqref{eqn:simplest_form}) now reads:
\begin{equation}
    \label{eqn:concept_with_z}
    \resizebox{.34\linewidth}{!}{$%
    \boldsymbol Y = (\boldsymbol Z \cdot \boldsymbol S ) \boldsymbol W_c^T 
    $}
\end{equation}
In this case, the output for each image is facilitated via the inner product computation between the weights of the linear layer and the \textit{effective concepts} as dictated by the introduced latent indicators $\boldsymbol{Z}$. 

A naive construction of the latent indicators $\boldsymbol z_i \in \{0,1\}^{M}, \ i=1, \dots, N$, for each example would result in: (i) significant computational overhead for storing each indicator, and (ii) no evident way to generalize the learned indicators to unseen examples. On this basis, we draw inspiration from recent Variational Bayesian advances \cite{chatzis2018indian, panousis2019nonparametric}  and postulate that the indicators $\boldsymbol Z$ are obtained via a data-driven random sampling procedure.  This translates to drawing samples from Bernoulli distributions, where the probability of concept presence is driven from a separate linear computation between the image embedding and a learnable weight matrix $\boldsymbol{W}_s \in \mathbb{R}^{K \times M}$, $K$ being the dimensionality of the embedding space; thus, for each input $\boldsymbol{X}_i$, this yields:
\begin{equation}
\setlength{\abovedisplayskip}{-2pt}
\label{eqn:bernoulli}
    \resizebox{.85\linewidth}{!}{$%
    q(\boldsymbol z_i) = \mathrm{Bernoulli}\Big( \boldsymbol z_i | \mathrm{sigmoid} \Big(E_I(\boldsymbol X_i) \boldsymbol W_s^T\Big) \Big)
    $}
\end{equation}
where the $\mathrm{sigmoid}(x) = 1/ (1 + e^{-x})$ nonlinearity is applied to convert the linear computation to probability. 

Thus, instead of only relying on the implicit relation between images and concepts learned using the distance of their embeddings, i.e., the similarity, and the linear layer, we exploit a separate source of information to devise an \textit{explicit} mechanism indicating concept presence in the context of the downstream task. This \textit{amortized} formulation bypasses both the aforementioned complications: we only need to store a single matrix with dimensions $K \times M$, while at the same time allowing for generalization to unseen examples. A graphical illustration of the envisioned architecture is depicted on Fig. \ref{fig:archs} (Right).

\paragraph{Training \& Prediction.} Assuming a dataset $\mathcal{D} = \{ (\boldsymbol{X}_i, \hat{Y}_i) \}_{i=1}^N$ and a concept set $\boldsymbol A =\{a_1, \dots, a_M\}$, the core training objective is the cross-entropy loss, denoted as $\mathrm{CE}(\hat{Y}_i, f(\boldsymbol X_i, \boldsymbol A))$, where $f(\boldsymbol X_i, \boldsymbol A)=\mathrm{Softmax}(\boldsymbol y_i)$ are the class probabilities; $\boldsymbol y_i$ is computed via Eqs.\eqref{eqn:simplest_form}, \eqref{eqn:concept_with_z}. Since the base model comprises only the weight matrix $\boldsymbol W_c$, it can be trained only via the cross entropy signal. 

When using the concept presence mechanism, the introduction of the binary latent indicators $\boldsymbol{Z}$, necessitates a different treatment of the training objective. In line with related literature \cite{panousis2019nonparametric}, we adopt the stochastic gradient Variational Bayes (SGVB) framework \cite{kingma2014autoencoding} for scalability. In this context, we impose an appropriate prior distribution for the latent indicators $\boldsymbol z_i$, i.e., a Bernoulli distribution, s.t., $p(\boldsymbol z_i) = \mathrm{Bernoulli}(\alpha), \ \forall i $, where $\alpha$ is a fixed non-negative constant. The resulting objective takes the form of an Evidence Lower Bound (ELBO) \cite{hoffman13a}:
\begin{equation}
    \label{eqn:elbo}
    \resizebox{.9\linewidth}{!}{$%
    \mathcal{L} = \sum_{i=1}^N \mathrm{CE}(\hat{Y}_i, f(\boldsymbol X_i, \boldsymbol A, \boldsymbol{z}_i)) - \beta \mathrm{KL}\left(q(\boldsymbol{z}_i) || p(\boldsymbol z_i) \right)
    $}
\end{equation}
where we augmented the notation to reflect the dependence on the binary indicators $\boldsymbol Z$ and $\beta$ is a scaling factor\cite{higgins2017betavae}. The second term is the Kullback-Leibler divergence; this encourages the posterior to be close to the prior. Thus, by setting $\alpha$ to a  value close to zero, we can effectively %introduce 
enforce a sparsity-inducing behavior in the learning process, while striking a balance between accuracy and sparsity without relying on any ad-hoc application or task specific thresholds. 

For training the model, we perform Monte Carlo sampling to estimate Eq. \eqref{eqn:elbo} using a single \textit{reparameterized} sample. Since the Bernoulli distribution is not amenable to the reparameterization trick\cite{kingma2014autoencoding}, we turn to its continuous relaxation \cite{jang, maddison} for training. During inference, we can directly draw samples from the trained posteriors $q(\boldsymbol Z)$ to investigate the per-example effect of a concept. For computing the per-class relevance, one can average the concept presence probability of all the examples of the class. 

At this point, it is important to highlight the flexibility of the proposed framework; instead of forcing an artificial sparsity on each class, we enable the model to learn the relevant concepts for each image in a data-driven manner, allowing for a varying number of activated concepts.

\section{Experimental Setup}

\begin{table*}[ht!]
    \centering
    \scalebox{0.8}{
    \begin{tabular}{lccccc}
        \hline
          & & \multicolumn{4}{c}{Dataset (Accuracy ($\%$) $\parallel$ Sparsity ($\%$)) }\\\cline{2-6}
         Model &  CIFAR10 & CIFAR100 & CUB200 & Places365 & ImageNet  \\\hline
         Standard \cite{labelfree}\textdagger & $88.80 $ & $70.10 $ & $76.70 $ & $48.56 $  &    $76.13 $      \\
         Standard (sparse) \cite{labelfree}\textdagger  & $82.96 $ & $58.34 $ & $75.96 $  & $38.46 $   & $74.35 $\\\hline
         Label-Free \cite{labelfree}\textdagger  & $86.37 $ & $65.27 $ & $74.59 $ & $43.71 $ & $71.98 $  \\\hline\hline
         CDM (RN50, w/o $\boldsymbol Z$) &  $81.90  \parallel --$ & $63.40  \parallel --$ & $64.70  \parallel --$  & $52.90  \parallel -- $   &    $71.20  \parallel --$\\
         CDM (RN50, w/ $\boldsymbol Z$)  &  $86.50  \parallel 2.55 $ & $67.60  \parallel 9.30$ & $72.26  \parallel 21.3 $ &$\mathbf{52.70}  \parallel 8.28$ & $72.20  \parallel 8.53$\\\hline
         CDM (ViT-B/16, w/o $\boldsymbol Z$) &  $94.45  \parallel --$ & $79.00  \parallel --$ & $75.10  \parallel --$ & $54.40  \parallel --$ & $77.90 \parallel --$ \\
         CDM (ViT-B/16, w/ $\boldsymbol Z$) & $\mathbf{95.30}  \parallel 1.69 $ & $\mathbf{80.50}  \parallel 3.38$ & $\mathbf{79.50}  \parallel 13.4 $ & $52.58  \parallel 8.00 $ & $\mathbf{79.30}  \parallel 6.96$ \\\hline
    \end{tabular}
    }
    \caption{Accuracy and Sparsity Results. By bold we note the best performing \textit{sparse} model. \textdagger indicates the reported performance. ``Standard'' models correspond to the non-interpretable backbones used in \cite{labelfree}.}
    \label{tab:perf}
\end{table*}

\begin{figure*}[h!]
    \centering
    \begin{subfigure}[b]{1.\textwidth}
        \centering
        \includegraphics[width = 0.8\textwidth]{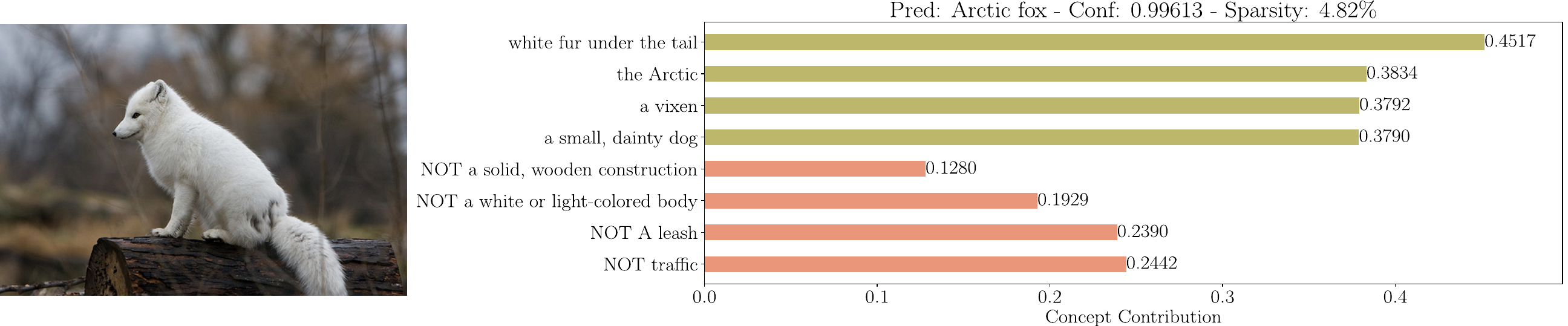}
    \end{subfigure}
    
    \begin{subfigure}[b]{1.\textwidth}
        \centering
        \includegraphics[width = 0.8\textwidth]{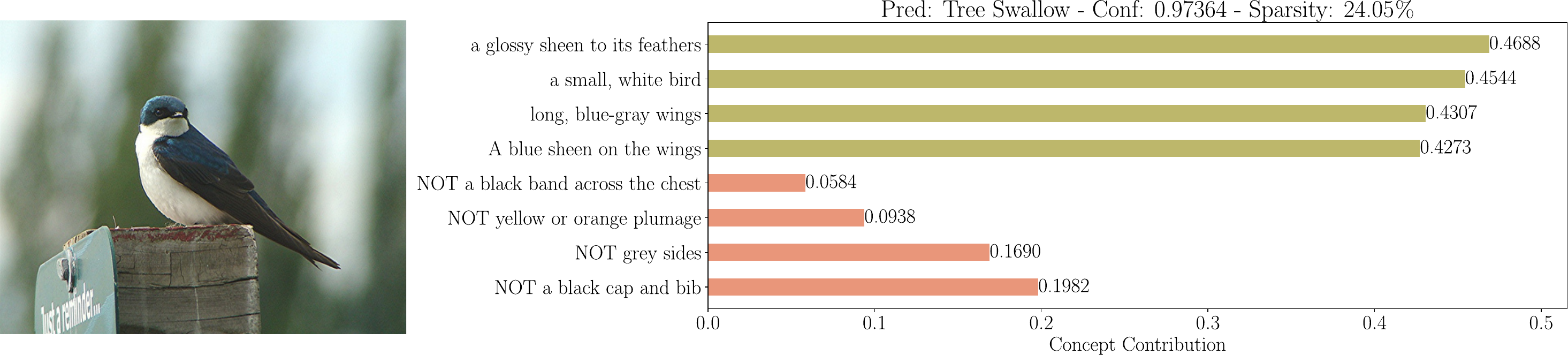}
    \end{subfigure}
    \caption{Concept Discovery investigation for an example from ImageNet-1k (Up) and CUB200 (Down). Khaki denotes positive contributions to the decision, while red negative.}
    \label{fig:visualizations}
    \vspace{-1em}
\end{figure*}

\paragraph{Datasets, Concepts Sets \& CLIP Backbones.} For thoroughly evaluating the proposed framework, we consider 5 datasets with varying characteristics: (i, ii) CIFAR10/100, (iii) CUB, (iv) Places365, and (v) ImageNet-1k. CIFAR-10/100 are standard recognition benchmarks, comprising $32 \times 32$ images, while CUB comprises higher resolution images focusing on fine-grained bird species identification. Their sizes also greatly vary, with CUB comprising $5900$ training examples, and Places365/ImageNet up to $1-2$ million. This highly diverse set of tasks will serve as an important showcase of the performance of the introduced mechanism. We consider the same concepts sets as in \cite{labelfree} for comparability, comprising 128, 824, 211, 2202 and 4505 concepts for CIFAR-10, CIFAR-100, CUB, Places365 and ImageNet, respectively. Finally, for the backbones of CLIP, we use the most common, i.e., ResNet50 (RN50) and ViT-B/16, which are frozen when computing the similarities. For our experiments, we set $\alpha = \beta = 10^{-4}$; we select the best performing learning rate among $\{10^{-3}, 5\cdot 10^{-3}, 10^{-2}\}$ for the linear layer. We set a higher learning rate for $\boldsymbol W_s$ ($10\times$) to facilitate learning of the discovery mechanism and train the models for a maximum of 2000 epochs for CIFAR10-100/CUB and 300 for ImageNet/Places365.

Our main competitor is Label Free-CBMs \cite{labelfree}. 
Even though it constitutes a highly different approach, in all settings, CLIP is used to compute similarities between images and concepts; then, instead of using them directly as in our framework, they are used to train a CBL. To achieve this, they use as backbones: (i) for CIFAR10/100, the CLIP RN50 encoder, (ii) for CUB, a CUB-trained RN18, and (iii) for Places/ImageNet, an ImageNet-trained RN50. After learning the projections, the GLM-SAGA solver \cite{wong2021} is used for learning a sparse linear layer, reporting $0.7-15\%$ non-zero weights without specifying the per class results. 

\paragraph{Results.} The obtained comparative results are depicted in Table \ref{tab:perf}. Therein, the sparsity values for our framework denotes the dataset-wise sparsity computed by averaging the per-example number of activated concepts over all samples. We observe that our concept discovery framework allows for extremely low concept retention while often 
it \textit{improves} the accuracy compared to the base model defined in Eq.\eqref{eqn:simplest_form}. Compared to related methods where sparsity is arbitrarily enforced on a class-wise level in an ad-hoc manner, in CDMs, the presence of a concept is inferred end-to-end on a data-driven per example basis. This facilitates balancing the trade-off between accuracy and sparsity during the learning process for each different example, greatly enhancing the flexibility of the framework. In Fig. \ref{fig:visualizations}, graphical illustrations of per-example discovered concepts are presented. The upper figure corresponds to an ImageNet \textit{test} sample and the lower from CUB. For the former, we observe a concept retention rate (sparsity) of $4.82\%$, translating to approximately $217$ active concepts out of the $4505$ potential concepts, with the top four most contributing being semantically similar to the example image. On the other hand, for the CUB example, we observe that the framework inferred a potentially necessary higher retention rate of $24.05\%$, translating to around $53$ out of $211$ active concepts, nevertheless exhibiting semantically similar most contributing concepts.

\section{Conclusions}

In this work, we proposed a novel framework towards intepretable networks based upon: (i) image-concept similarities arising from CLIP models, (ii) a linear layer for classification, and (iii) a novel data-driven mechanism for per-example concept discovery. The experimental results vouch for the efficacy of the approach. Our CDMs retain or even improve classification performance, while at the same time enabling very high per-example sparsity without limiting the flexibility of the concept selection mechanism.

{\small
\bibliographystyle{ieee_fullname}
\bibliography{egbib}
}

\newpage
\appendix

\section{Limitations \& Future Work}
One limitation of the proposed framework is the dependence on a CLIP-like backbone to obtain the images-concepts similarities. On this basis: (i) there is no ``easy'' way to recover from the backbone's \textit{concept omissions}. Indeed, if the image-text model assigns a large similarity value to a particular unrelated concept, this can be removed via the concept discovery mechanism. However, if the backbone assigns zero similarity between an image and a given concept, despite the latter being present in the image, it will not contribute to the downstream task. (ii) The results depend on the suitability of the backbone to the considered application; thus, if the backbone can not adequately model the underlying data due to either its architecture or concepts missing from (or biases contained in)  the data used for pretraining, the final performance will reflect that, even if the introduced CDM framework somewhat alleviates this issue via the concept discovery mechanism. In this context, even though the experimental results suggest that using the ViT-B CLIP backbone can yield significant performance, it may not work in all cases. However, the proposed framework constitutes a general proposal: any future advances on multi-modal models can be easily incorporated by changing the projection backbone. In our future work, we aim to lessen the dependence on the pretrained backbones and find ways to either adjust the arising similarities or combine different or multiple image and text encoders to match the downstream task.

\section{Bernoulli Relaxation \& Inference}

\paragraph{Training.} As already noted in the main text, to estimate the ELBO in Eq.\eqref{eqn:elbo}, we perform Monte-Carlo sampling, with a single reparameterized sample. However, the Bernoulli distribution is not amenable to the reparameterization trick \cite{kingma2014autoencoding}. To this end, we resort to its continuous relaxation\cite{maddison,jang}. 

Let us denote by $\boldsymbol{\tilde{z}_i}$, the probabilities of $q(\boldsymbol z_i) , \ i=1, \dots N$. We can directly draw reparameterized samples $\boldsymbol{\hat{z}_i} \in (0,1)^M$ from the continuous relaxation as:
\begin{align}
    \boldsymbol{\hat{z}_i} = \frac{1}{1 + \exp \left( - (\log \boldsymbol{\tilde{z}_i} + L) / \tau \right) }
    \label{eqn:bernoulli_relaxed}
\end{align}
where $L\in\mathbb{R}$ denotes samples from the Logistic function, such that:
\begin{align}
    L = \log U - \log ( 1-U), \quad U \sim \mathrm{Uniform}(0,1)
\end{align}
where $\tau$ is called the \textit{tempetature} parameter; this controls the degree of the approximation: the higher the value the more uniform the produced samples and vice versa. We set $\tau$ to $0.1$ in all the experimental evaluations.

\paragraph{Inference.} During inference, and for each test example, we draw sample(s) from the Bernoulli distribution defined in Eq. \eqref{eqn:bernoulli} to obtain the binary indicator vector $\boldsymbol{z} \in \{0, 1\}^M$: each entry therein denotes the presence or absence of a concept for the given example. This is used to: (i) compute the output of the network according to Eq. \eqref{eqn:concept_with_z} and subsequently the loss function (in our case the cross-entropy), and (ii) examine each concept activated for the given example.

\section{Ablation Study.}
For learning the auxiliary binary latent variables $\boldsymbol Z$, we introduced appropriate prior and posterior distributions and constructed the ELBO. In this context, we introduced two additional hyperparameters: (i) the prior parameter $\alpha$ and (ii) the scale of the KL divergence, $
\beta$. Here, we examine the effect of these parameters of the final performance using the ViT-B/16 backbone and the CUB dataset and two different learning rates $10^{-2}$ and $10^{-3}$. In Table \ref{tab:ablation}, we report the performance of the framework in terms of accuracy and sparsity for different values of $\alpha, \beta$. 

\begin{table}[h!]
    \centering
    \begin{tabular}{cc|cc}
        \hline
         $\alpha$ & $\beta$  & Accuracy (\%) & Sparsity (\%) \\\hline
         $10^{-2}$ & $10^{-4}$  & $80.67$ & $23.38$\\ 
         $10^{-4}$ & $10^{-4}$  & $80.00$ & $16.12$\\
         $10^{-4}$ & $10^{-5}$  & $79.70$ & $14.07$\\\hline
         $10^{-3}$ & $10^{-4}$  & $82.23$ & $37.7$\\
         $10^{-3}$ & $5\cdot 10^{-4}$  & $81.40$ & $20.93$\\
         $5\cdot 10^{-4}$ & $10^{-3}$  & $81.07$ & $17.61$\\\hline
    \end{tabular}
    \caption{Ablation results on the impact of the hyperparameters on: (i) the resulting accuracy and (ii) the emerging sparsity using the ViT-B/16 backbone for CLIP and CUB200 as the training dataset. The learning rate in this study was set to $5 \cdot 10^{-3}$ for the top table and $10^{-3}$ for the bottom table respectively.}
    \label{tab:ablation}
\end{table}

\end{document}